\documentclass[11pt]{article}
\usepackage[final]{acl}
\usepackage{times}
\usepackage{latexsym}
\usepackage[T1]{fontenc}
\usepackage[utf8]{inputenc}
\usepackage{microtype}
\usepackage{inconsolata}
\usepackage{graphicx}
\usepackage{booktabs}
\usepackage{multirow}
\usepackage{colortbl}
\usepackage{amsmath}
\usepackage{amssymb}
\usepackage{float}
\usepackage{stfloats}
\usepackage{placeins}

\newcommand{\highlightrow}{\rowcolor[gray]{0.94}}

\title{LLM4LLM: Bridging Kernel Benchmarks and Real Deployment via Closed-Loop Agentic Optimization}

\author{
 \textbf{Hui Zeng\textsuperscript{1,2}},
 \textbf{Pengfei Yang\textsuperscript{1,4,*}},
 \textbf{Yanxin Chen\textsuperscript{1}},
 \textbf{Fusong Ju\textsuperscript{2}},
 \textbf{Xinran Wei\textsuperscript{2,3,*}}
\\
\\
 \textsuperscript{1}Xidian University,
 \textsuperscript{2}Zhongguancun Academy,
 \textsuperscript{3}Zhongguancun Institute of Artificial Intelligence
\\
 \textsuperscript{4}National Key Laboratory of Advanced Communication Networks, Shijiazhuang, Hebei, P. R. China
\\
 \small{
   \textsuperscript{*}\textbf{Corresponding authors:}
   \href{mailto:pfyang@xidian.edu.cn}{pfyang@xidian.edu.cn},
   \href{mailto:weixinran@zgci.ac.cn}{weixinran@zgci.ac.cn}
 }
}

\begin{document}
\maketitle

\begin{abstract}
Large language models have become increasingly capable agents for low-level code and kernel optimization, but isolated kernel benchmarks provide only a proxy for the deployment behavior that matters in language-model inference.
We identify a benchmark-to-deployment gap: candidate kernels that appear correct and fast in standalone harnesses can exhibit different performance, safety, or phase behavior after integration into a real inference workload.
We introduce LLM4LLM, a deployment-aware closed-loop optimization framework that starts from a target inference script, extracts phase-aware optimization tasks, searches with an experience-guided episodic agent, and accepts patches through in-model validation.
Across ten language-model inference workloads on A100 and H100 GPUs, LLM4LLM improves end-to-end latency for every evaluated model, achieving 3.91$\times$/6.98$\times$ geometric-mean speedups on A100/H100; as supporting kernel-level evidence, it also attains up to 2.745$\times$ GeoMean speedup on KernelBench Level 2.
\end{abstract}

\section{Introduction}
\label{sec:intro}

Recent progress in LLM-driven code generation has made automated kernel and program optimization a credible direction for performance engineering.
Code-specialized models support synthesis, infilling, and editing \citep{chen2021evaluating,wang2021codet5}.
Competitive-programming and execution-feedback systems extend this ability to search-guided generation \citep{li2022competition,le2022coderl}.
Iterative agents improve programs through feedback and debugging traces \citep{madaan2023self,chen2024selfdebug}.
Recent kernel benchmarks further show that LLMs can generate low-level GPU programs \citep{ouyang2025kernelbench}.
Most existing evaluations, however, are centered on isolated kernels or extracted program harnesses: a candidate is compiled, executed on synthetic inputs, checked against a reference implementation, and ranked by standalone latency.
For language-model inference, this measures an intermediate signal.
The deployment objective is the behavior of the patched model under the target workload, including phase semantics, cache state, dispatch overhead, memory residency, and end-to-end latency.

We study the mismatch between benchmark-level optimization and model-level deployment.
Autoregressive inference exposes this mismatch clearly.
Prefill and decode have different shapes, cache states, memory-access patterns, and latency sensitivity; a candidate optimized for one phase can transfer unevenly to the other.
Integration can also change the execution environment through shape guards, dispatch logic, allocator state, and interactions with existing high-performance kernels.
We refer to this prediction mismatch as the \emph{benchmark-to-deployment gap}.
The gap appears as speedup reversal, deployment-time runtime failure, and phase-specific behavior that is absent from isolated qualification.

LLM4LLM\footnote{Code is available at \url{https://github.com/hzeng2000/LLM4LLM}.} addresses this gap by making deployment context part of the optimization loop.
Starting from a user-provided inference script, the system profiles real workloads and extracts optimization tasks at deployable module boundaries.
It constructs phase-aware tasks for prefill and decode, searches for candidates through experience-guided episodic optimization, and accepts patches only after in-model validation.
The episodic agent compresses useful experience from earlier attempts into concise constraints, allowing search to reuse successful patterns while discarding stale context.
The resulting loop ties profiling, generation, verification, acceptance, and deployment into a single optimization process.

Our contributions are:
\begin{itemize}
    \item We characterize the benchmark-to-deployment gap for LLM-based kernel optimization through a taxonomy of failure modes, showing how isolated qualification can diverge from deployment-time acceptance.
    \item We introduce LLM4LLM, a closed-loop agentic optimization framework that combines real-model profiling, phase-aware task formulation, experience-guided episodic search, and in-model acceptance.
    \item We evaluate LLM4LLM on real inference workloads and KernelBench Level 2, showing end-to-end gains across transformer, state-space, and recurrent language-model families with 3.91$\times$/6.98$\times$ geometric-mean speedups on A100/H100, and GeoMean kernel speedups of 2.745$\times$/2.628$\times$ on KernelBench Level 2.
\end{itemize}

\begin{figure}[H]
  \centering
  \makebox[\columnwidth][c]{%
    \includegraphics[width=1.04\columnwidth,trim=18pt 8pt 12pt 16pt,clip]{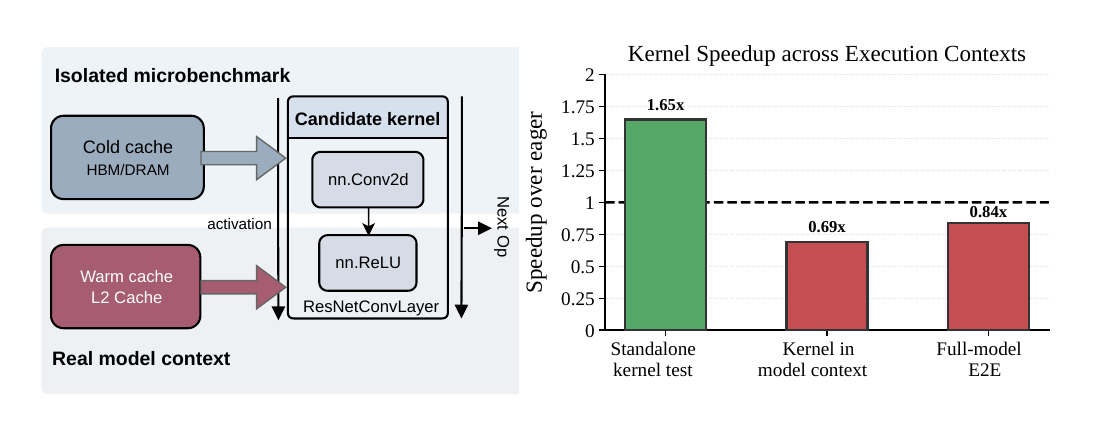}%
  }
  \caption{Divergent isolated and deployment outcomes for an optimized \texttt{ResNetConvLayer}. The left panel contrasts standalone cold-cache benchmarking with in-model warm-context execution for the same candidate kernel. The right panel reports speedup from standalone kernel testing, the same kernel measured inside the model context, and the resulting full-model end-to-end execution.}
  \label{fig:motivating-example}
\end{figure}

\section{The Benchmark-to-Deployment Gap}
\label{sec:gap}

\subsection{Motivating Evidence}
\label{sec:gap-cases}

We begin with a representative case that isolates the effect of execution context, shown in Figure~\ref{fig:motivating-example}.
A convolution layer extracted from a standard vision backbone is optimized by an LLM-based agent and evaluated in two settings.
In the isolated setting, the candidate is compiled as a standalone module, invoked on freshly allocated random inputs, and timed independently.
In the deployment setting, the same candidate is patched into the full model and evaluated in the target inference workload.
The isolated benchmark reports a substantial speedup, while the integrated model becomes slower end to end.

The mechanism is memory residency.
In the standalone benchmark, the operator pays the cost of loading cold inputs from memory.
Inside the model, the same operator consumes activations produced immediately by the preceding layer, so the baseline benefits from warm-cache execution and producer-consumer locality.
A memory-access pattern that improves cold standalone timing can lose its advantage once the surrounding model supplies different cache state and scheduling context.
This case shows how proxy measurements can misrank candidates before deployment.

The same gap also appears in forms that are specific to deployability and autoregressive inference.
First, a candidate may pass isolated correctness while carrying a latent memory-safety error.
Standalone tensors are often allocated with adjacent mapped memory, so an out-of-bounds read can return a value without faulting.
The allocator state of a full model can place the same access in a faulting region, producing a deployment-time CUDA error.
Second, phase specialization can make a kernel-level gain local to only one part of generation.
For example, a candidate that optimizes attention during prefill may be applicable when \texttt{q\_len > 1}, while decode attention runs with \texttt{q\_len = 1} and a populated KV cache.
The prefill fast path can therefore fall back, or provide no benefit, in the decode phase.

\begin{figure*}[!t]
  \centering
  \includegraphics[width=0.95\textwidth]{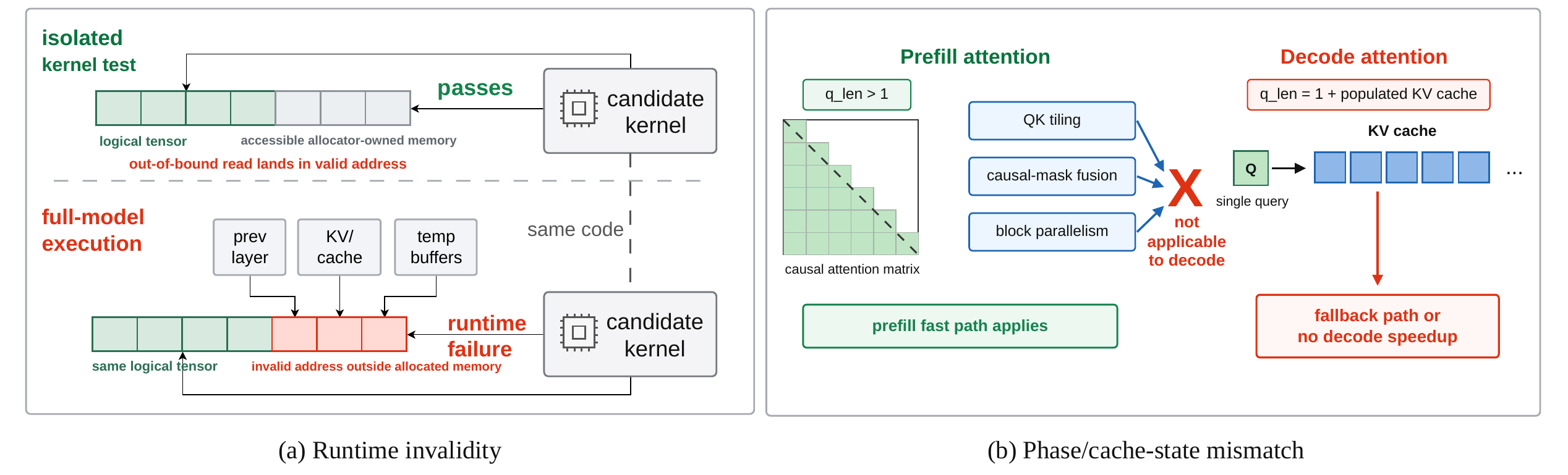}
  \caption{Deployment-specific manifestations of the benchmark-to-deployment gap. The left panel shows how a latent out-of-bounds access can pass isolated qualification under a fresh allocator state yet fail after integration into the full model runtime. The right panel shows that an attention candidate specialized for prefill can be inapplicable to decode attention, so a kernel-level gain does not necessarily transfer across generation phases.}
  \label{fig:benchmark-to-deployment}
\end{figure*}

These cases establish three deployment concerns: performance transfer, runtime validity, and phase-correct behavior.
They also motivate a distinction between isolated qualification and deployment-time acceptance, as illustrated in Figure~\ref{fig:benchmark-to-deployment}.

\subsection{Formulation and Implications}
\label{sec:gap-formulation}

We distinguish four evaluation stages.
\emph{Isolated qualification} checks numerical correctness and latency in a standalone harness.
\emph{Search-time validation} strengthens the proxy by adding real inputs, runtime-safety gates, or in-context timing.
\emph{Deployment-time acceptance} evaluates a candidate after it has been inserted into the target model instance.
\emph{Final end-to-end evaluation} measures the patched model under the full workload.

The benchmark-to-deployment gap is the mismatch between isolated qualification and deployment-time acceptance.
Execution context, phase behavior, and integration constraints all contribute to this mismatch.
Isolated benchmarks remain valuable because they allow rapid candidate generation and screening, while deployment-time acceptance determines whether a candidate improves the target model.
This leads to a methodological requirement: candidate generation, validation, and patch acceptance form a closed loop around the real workload.

\section{Method}
\label{sec:method}

\begin{figure*}[!t]
  \centering
  \includegraphics[width=0.98\textwidth]{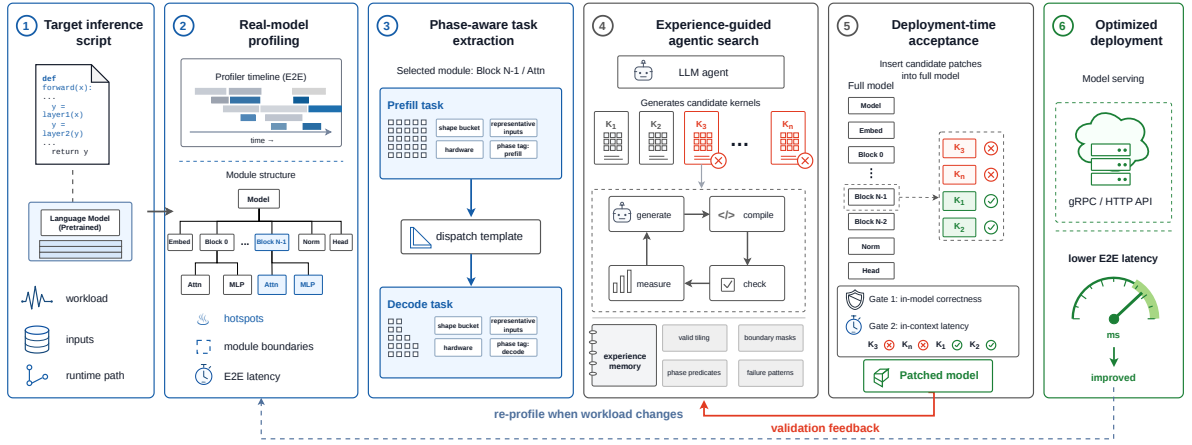}
  \caption{Overview of LLM4LLM. The framework starts from a target inference script, identifies deployable hotspots through real-model profiling, constructs phase-aware tasks, searches for candidate kernels with experience-guided agentic optimization, and accepts patches only after in-model correctness and latency validation. Validation feedback is returned to the search loop, while accepted patches produce the optimized deployment.}
  \label{fig:method-overview}
\end{figure*}

Figure~\ref{fig:method-overview} summarizes the LLM4LLM optimization loop.
Starting from a target inference script, the system profiles the real workload, extracts phase-aware optimization tasks, searches with an experience-guided agent, and accepts patches through model-integrated validation.
We denote the target model instance by \(M\), the deployment workload induced by the inference script by \(\mathcal{W}\), and the set of autoregressive phases by \(\mathcal{P}\).
Here \(M\) is a concrete model instance, while \(\mathcal{W}\) and \(\mathcal{P}\) are structured collections.
For a module instance \(m\) and phase \(p\in\mathcal{P}\), profiling records the module time \(t(m,p)\) and the total phase latency \(T(p)\).
LLM4LLM selects deployable optimization units using the workload-weighted hotspot score
\begin{equation}
s(m)=\sum_{p\in\mathcal{P}}\omega_p \frac{t(m,p)}{T(p)},
\label{eq:hotspot-score}
\end{equation}
where \(\omega_p\) is determined by the measured phase frequency or by the evaluation workload.
The score is used for task extraction, and final acceptance is made inside the target model.

We next detail the three stages of this loop: phase-aware task extraction, experience-guided search, and deployment-time acceptance.

\subsection{Hotspot Discovery and Phase-Aware Extraction}
\label{sec:method-task}

LLM4LLM starts from the inference script supplied by the user.
The script defines the workload, input regime, runtime path, and performance objective.
The system profiles this execution hierarchically and selects semantic module instances whose replacement can affect end-to-end latency.
A module is considered deployable when it has a stable call boundary, reproducible input and output tensors, and a fallback implementation for unoptimized regimes.

For each selected module, LLM4LLM serializes an optimization task containing representative inputs, output references, shape information, module-family metadata, hardware scope, and phase tags.
For autoregressive inference, prefill and decode are represented as separate tasks when their execution semantics diverge.
When both phases are optimized, extraction also records a dispatch template that later reassembles phase-specialized candidates into a single patched module.
The task preserves the deployment facts that affect validity, including tensor layout, cache state, phase predicate, and observed shape regimes.

\subsection{Experience-Guided Episodic Optimization}
\label{sec:method-optimization}

The optimizer searches through repeated generate-verify-decide episodes.
Within an episode, the agent proposes candidate kernels, compiles them, runs correctness checks, measures latency, and inspects failures.
Each episode therefore produces a concrete validation trace: candidate code, compiler diagnostics, numerical errors, runtime failures, and latency measurements.
LLM4LLM uses this trace to decide whether the current search context is still productive or has converged to local repair of a narrow failure mode.
For a candidate \(c\) on task \(\tau\), search-time validation assigns the utility
\begin{equation}
u(c,\tau)=\frac{t_{\mathrm{ref}}(\tau)}{t_c(\tau)}
\mathbf{1}\{\mathrm{err}(c,\tau)\le\epsilon,\ \mathrm{safe}(c,\tau)\},
\label{eq:search-utility}
\end{equation}
where \(t_{\mathrm{ref}}\) is the reference task latency, \(t_c\) is the candidate latency, \(\mathbf{1}\{\cdot\}\) is the indicator function, \(\mathrm{err}\) is the numerical error against recorded outputs, and \(\mathrm{safe}\) denotes compilation and runtime-safety checks.
This utility ranks promising candidates within the extracted task; deployment-time acceptance remains the final decision.

At episode boundaries, the system distills the validation trace into compact experience.
The distilled record contains durable constraints and search evidence, including valid tiling choices, boundary-mask requirements, phase predicates, numerical constraints, failure signatures, and the best observed performance regime.
The next episode is initialized with the original task specification and this experience record; the full turn-by-turn transcript is archived.
The summary for episode \(r\) is a bounded record \(E_r\) that stores accepted constraints, rejected failure modes, and the best candidate family observed so far.
Restarting from the task plus \(E_r\) preserves useful evidence while reducing the influence of long local repair trajectories.

\begin{figure}[t]
  \centering
  \includegraphics[width=\columnwidth]{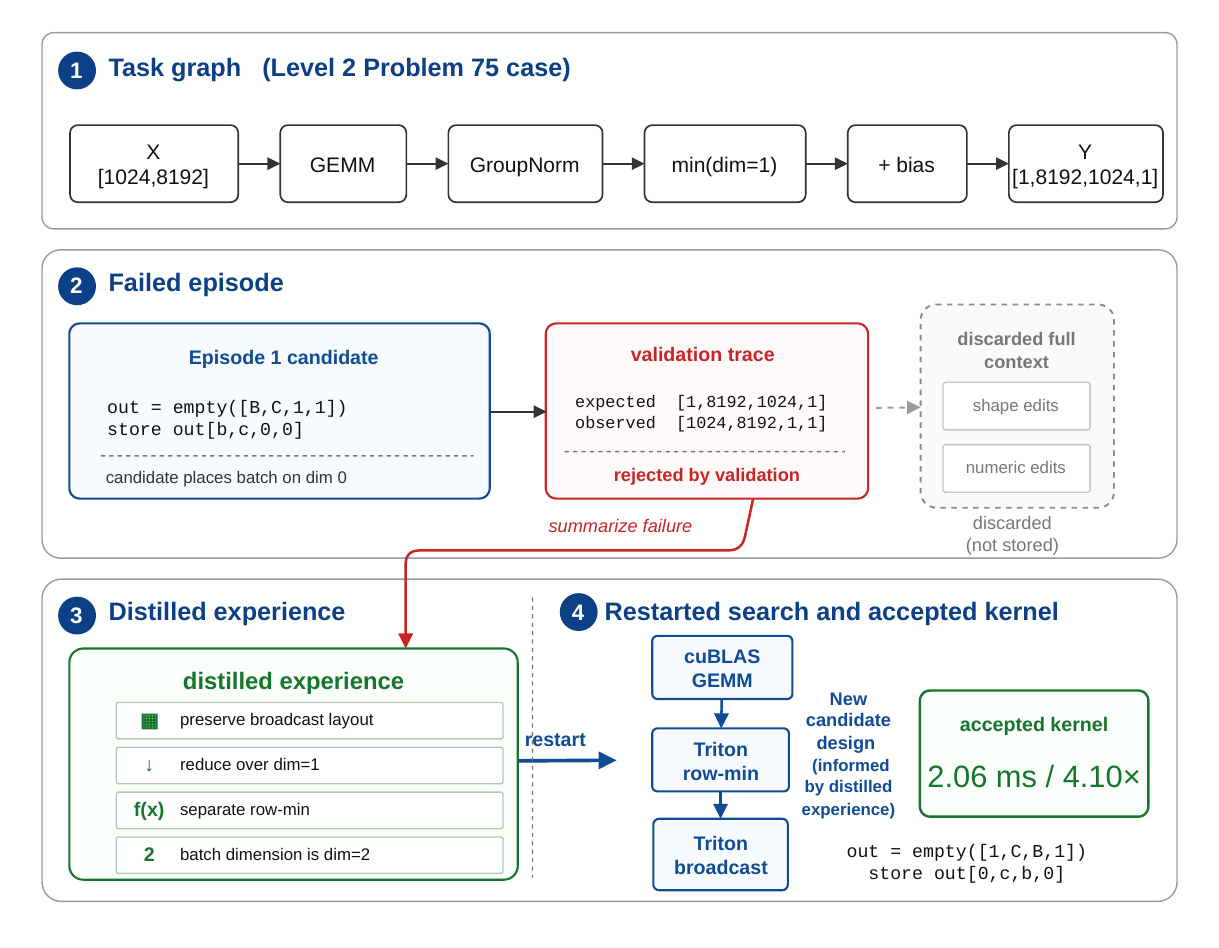}
  \caption{Experience-guided episodic optimization. Candidate generation and validation form a bounded episode. At the episode boundary, LLM4LLM summarizes the useful validation evidence into compact experience, archives the transient repair history, and restarts search from the task specification plus the distilled record.}
  \label{fig:episodic-optimization}
\end{figure}
Figure~\ref{fig:episodic-optimization} illustrates this mechanism on a representative optimization episode. This design treats validation feedback as reusable optimization evidence.
Episode-level memory captures task-specific lessons, while family-scoped memory captures patterns that transfer across related modules, shape buckets, and hardware targets.
The search process remains grounded in observed verification constraints while allowing later episodes to explore implementation structures beyond earlier local edits.

\subsection{Deployment-Time Acceptance and Patching}
\label{sec:method-acceptance}

Search produces qualified candidates; deployment-time acceptance decides which candidates enter the model.
LLM4LLM ranks candidates by module compatibility, shape coverage, phase compatibility, and predicted impact.
Each candidate is inserted into the target model instance and evaluated for in-context correctness and latency.
Accepted candidates become patches; rejected candidates provide feedback for subsequent search.
Let \(\ell(c)=L(M[c],\mathcal{W})/L(M,\mathcal{W})\) be normalized deployment latency, and let \(\mathcal{C}_{\mathrm{dep}}=\{c\in\mathcal{C}:\mathrm{ok}_{\mathrm{dep}}(c)\}\) be the candidates passing in-model correctness, runtime compatibility, and shape-guard coverage.
Deployment-time acceptance selects
\begin{equation}
c^\star=\arg\min_{c\in\mathcal{C}_{\mathrm{dep}}}\ell(c),
\qquad \ell(c^\star)\le 1-\delta .
\label{eq:deployment-acceptance}
\end{equation}
where \(\delta\) is the minimum deployment improvement required to accept a patch.
Equation~\ref{eq:deployment-acceptance} ties candidate generation to the evaluation objective.

For phase-specialized replacements, the patched module uses the dispatch template generated during extraction.
Runtime guards preserve shape compatibility, and fallback paths preserve execution for unoptimized regimes.
The final patch therefore reflects both the agent's generated implementation and the deployment constraints of the target workload.

\begin{figure*}[!t]
  \centering
  \includegraphics[width=0.80\textwidth]{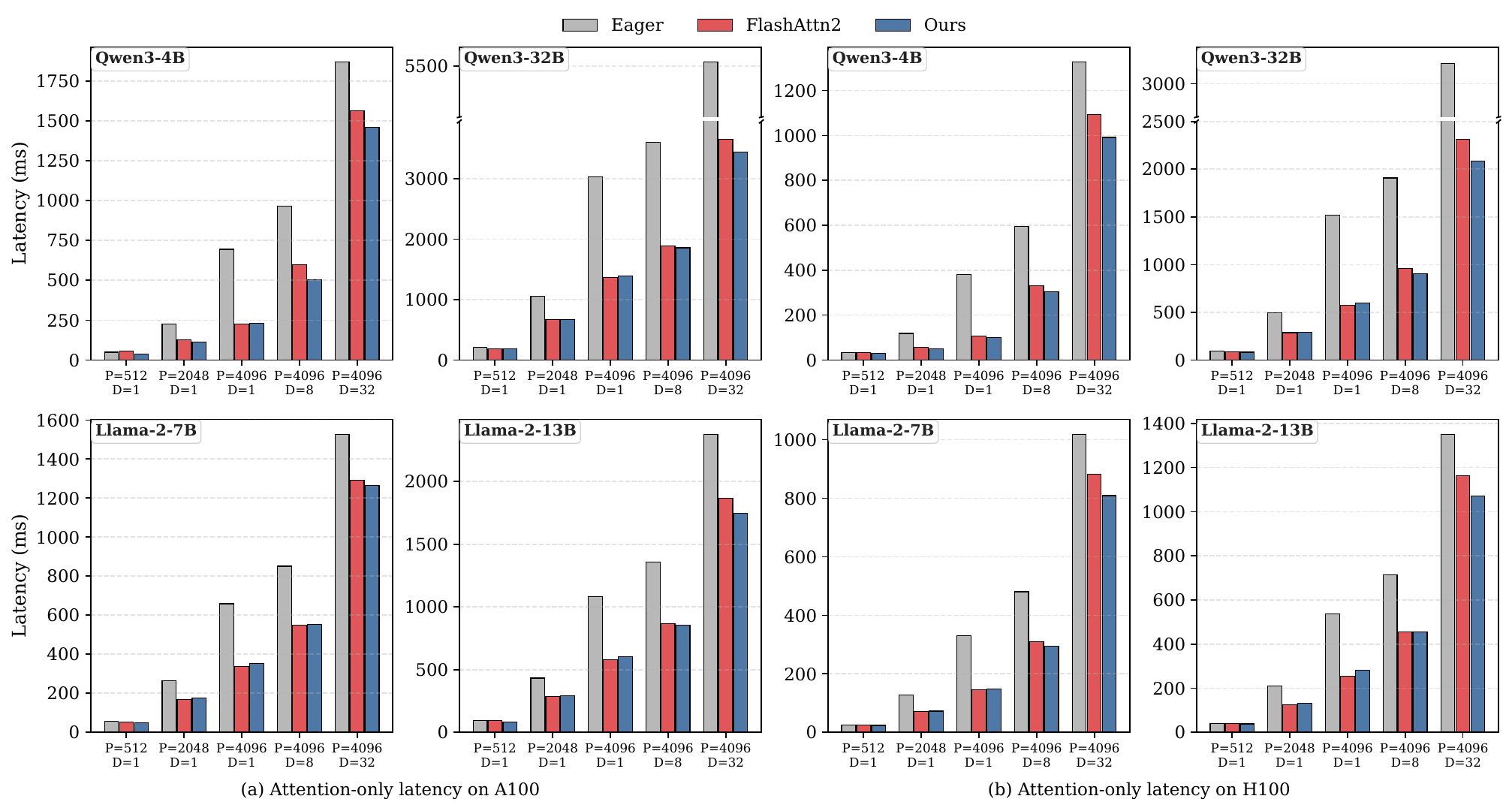}
  \caption{Scope-matched comparison on transformer-family attention workloads. Each panel reports latency for eager execution, LLM4LLM, and expert deployment kernels across representative workload settings on A100 and H100. Lower is better.}
  \label{fig:attention-scope-compare}
\end{figure*}

\section{Evaluation}
\label{sec:eval}

\subsection{Experimental Setup}
\label{sec:eval-setup}

We evaluate LLM4LLM on language-model inference workloads spanning transformer-family models, Mamba-family state-space models, and RecurrentGemma.
Experiments are conducted on A100 and H100 GPUs.
The primary metric is end-to-end latency of the target inference workload, reported together with speedup over eager execution.
LLM4LLM starts optimization from the eager PyTorch execution path.
For scope-matched comparisons, we evaluate attention-only and mixer-only settings against strong deployment baselines, including FlashAttention for attention and Mamba fast paths for state-space mixers.
Here, Mamba fast paths refer to hand-optimized kernels used by the Mamba and Mamba-2 implementations: \texttt{mamba\_ssm} for selective-scan and state-space mixer execution, and \texttt{causal-conv1d} for causal depthwise convolution \citep{gu2023mamba,mamba2,dao2024causalconv1d}.
All reported patches pass model-integrated correctness checks before latency measurement.
The evaluation follows the same acceptance path used by the method: candidates are validated on extracted tasks, inserted into the model instance, and measured through the target inference script.
We report latency because the objective includes launch overhead, guards, cache behavior, and interactions with surrounding model code.

\subsection{End-to-End Results on Language-Model Families}
\label{sec:eval-main}

\begin{table}[H]
\centering
\small
\setlength{\tabcolsep}{5pt}
\renewcommand{\arraystretch}{1.08}
\caption{End-to-end latency (ms) across language-model families on A100 and H100. Lower latency and higher speedup are better.}
\label{tab:main-e2e}
\resizebox{\columnwidth}{!}{%
\begin{tabular}{lrrrrrr}
\toprule
\multirow{2}{*}{\textbf{Model}} & \multicolumn{3}{c}{\textbf{A100}} & \multicolumn{3}{c}{\textbf{H100}} \\
\cmidrule(lr){2-4}\cmidrule(lr){5-7}
 & Eager & Ours & Speedup & Eager & Ours & Speedup \\
\midrule
Qwen3-4B & 693.9 & 229.0 & 3.03$\times$ & 380.2 & 100.7 & 3.78$\times$ \\
Qwen3-32B & 3032.3 & 1387.1 & 2.19$\times$ & 1516.7 & 594.4 & 2.55$\times$ \\
Llama-2-7B & 658.8 & 353.9 & 1.86$\times$ & 329.8 & 148.4 & 2.22$\times$ \\
Llama-2-13B & 1082.2 & 603.7 & 1.79$\times$ & 538.1 & 280.7 & 1.92$\times$ \\
Mamba (130M) & 626.4 & 28.0 & 22.37$\times$ & 537.7 & 16.4 & 32.79$\times$ \\
Mamba (2.8B) & 959.2 & 118.5 & 8.09$\times$ & 1446.4 & 50.5 & 28.64$\times$ \\
Mamba2 (130M) & 198.5 & 33.0 & 6.02$\times$ & 137.8 & 22.0 & 6.26$\times$ \\
Mamba2 (2.7B) & 665.4 & 43.7 & 15.23$\times$ & 445.0 & 31.7 & 14.04$\times$ \\
RecurrentGemma-2B & 329.4 & 180.2 & 1.83$\times$ & 287.9 & 26.8 & 10.74$\times$ \\
RecurrentGemma-9B & 712.5 & 575.5 & 1.24$\times$ & 409.8 & 54.3 & 7.55$\times$ \\
\bottomrule
\end{tabular}
}
\end{table}

Table~\ref{tab:main-e2e} shows that the closed-loop optimization produces end-to-end gains across all evaluated families.
These numbers are measured after patch insertion, so they reflect the realized effect of candidate kernels together with dispatch overhead, shape guards, cache state, and surrounding model code.
The transformer-family results show steady improvements on both GPUs.
For these models, profiling usually selects attention-dominated regions and adjacent tensor operations whose cost remains visible after existing fused attention paths are enabled.
The gains therefore reflect deployment-level replacement boundaries and model-integrated acceptance.

The state-space and recurrent families show a different pattern.
Mamba-family workloads concentrate latency in mixer modules whose computation is regular enough for generated Triton kernels to cover a large share of the inference path.
RecurrentGemma exposes another profile: the dominant recurring module is \texttt{RecurrentGemmaRglru}, whose cost is tied to gated recurrent updates, state movement, and phase-dependent memory behavior.
This observation motivates the first profiling stage of LLM4LLM.
The dominant deployable kernel is model-family dependent: transformer traces emphasize attention regions, Mamba traces emphasize state-space mixers, and RecurrentGemma traces emphasize the RGLRU recurrent core.
A fixed attention-first policy allocates budget poorly for these families; profiling-first extraction directs search to the modules that dominate the actual workload.
The A100--H100 differences further show that the accepted patch depends on both model structure and hardware execution context.

\subsection{Comparisons with Strong Deployment Baselines}
\label{sec:eval-baselines}

\begin{figure*}[!t]
  \centering
  \includegraphics[width=0.80\textwidth]{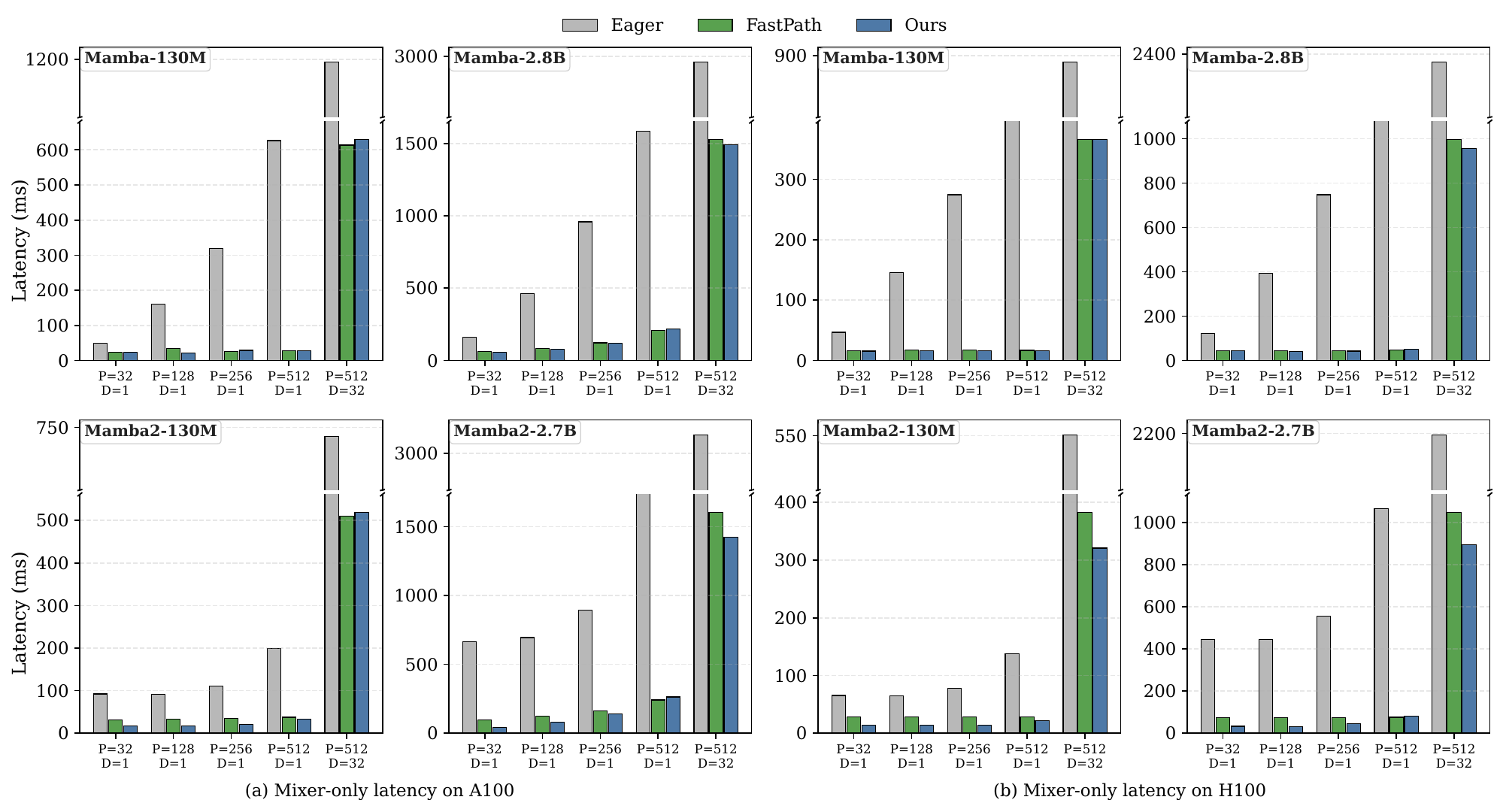}
  \caption{Scope-matched comparison on state-space mixer workloads. Each panel reports latency for eager execution, LLM4LLM, and Mamba-family fast paths based on \texttt{mamba\_ssm} and \texttt{causal-conv1d} across representative workload settings on A100 and H100. Lower is better.}
  \label{fig:mamba-scope-compare}
\end{figure*}

Figures~\ref{fig:attention-scope-compare} and~\ref{fig:mamba-scope-compare} compare LLM4LLM against strong deployment baselines under matched optimization scope.
The attention comparison evaluates the setting where expert kernels are especially mature.
LLM4LLM reaches competitive latency in several prompt/decode regimes, while the remaining gaps identify workload shapes where specialized attention implementations retain an advantage.
This result is useful for deployment because it separates two questions: whether an LLM-generated candidate can execute correctly inside the model, and whether the measured attention region is the best use of the search budget for that model.

The mixer comparison gives complementary evidence on a family with a different dominant operation.
When profiling selects state-space mixer boundaries, generated replacements can absorb surrounding reshapes, projections, and elementwise updates that are outside the scope of a single vendor or library kernel.
This wider deployable boundary explains why LLM4LLM can approach or improve over the hand-optimized \texttt{mamba\_ssm} and \texttt{causal-conv1d} paths in several regimes.
Together with the RecurrentGemma results in Table~\ref{tab:main-e2e}, the scope-matched figures support a profiling-driven view of optimization: attention, state-space mixers, and recurrent RGLRU kernels each become the right target only when they dominate the measured workload.
The deployment baselines therefore serve as strong references for their own scopes, while the closed-loop acceptance step determines which generated patch improves the full inference path.

\begin{table*}[!t]
\centering
\caption{KernelBench Level-2 comparison on A100 and H100 GPUs. Pass, Fast1, and Fast2 are reported as percentages. Mean and GeoMean are speedups over PyTorch eager unless otherwise specified.}
\label{tab:kernelbench-comparison}
\scriptsize
\setlength{\tabcolsep}{2.0pt}
\renewcommand{\arraystretch}{1.0}
\resizebox{\textwidth}{!}{%
\begin{tabular}{@{}lcccccc@{\hspace{0.8em}}lcccccc@{}}
\toprule
\multicolumn{7}{c}{\textbf{A100-SXM4-80GB}} &
\multicolumn{7}{c}{\textbf{H100-80GB-HBM3}} \\
\cmidrule(lr){1-7}\cmidrule(lr){8-14}
\textbf{Method} & \textbf{Lang.} & \textbf{Pass} & \textbf{Fast1} & \textbf{Fast2} & \textbf{Mean} & \textbf{GeoMean} &
\textbf{Method} & \textbf{Lang.} & \textbf{Pass} & \textbf{Fast1} & \textbf{Fast2} & \textbf{Mean} & \textbf{GeoMean} \\
\midrule
CUDA-L1 & CUDA & 100 & 98 & -- & 3.55 & -- &
CUDA-L1 & CUDA & 96 & 96 & -- & 6.64 & -- \\
KernelSkill & CUDA & 100 & 100 & -- & 2.82 & -- &
KernelBlaster & CUDA & 81 & 84.85 & -- & 10.223 & 2.592 \\
STARK (A100-40G) & CUDA & 100 & 100 & -- & 2.69 & -- &
QiMeng-Kernel & Triton & 99 & 86 & 12 & 1.28 & -- \\
QiMeng-Kernel & Triton & 99 & 66 & 8 & 1.22 & -- &
AI CUDA Engineer & CUDA & 82 & 61.33 & -- & 1.356 & 1.214 \\
\highlightrow LLM4LLM & Triton & 100 & 97 & 39 & 22.157 & 2.745 &
LLM4LLM & Triton & 99 & 95 & 41 & 15.847 & 2.628 \\
\bottomrule
\end{tabular}}
\end{table*}

\subsection{KernelBench Comparison}
\label{sec:eval-kernelbench}

KernelBench provides supporting evidence for kernel-level capability and agent search behavior while the preceding experiments measure deployment transfer.

We first compare LLM4LLM with recent LLM-based kernel optimization systems on KernelBench Level 2.
The comparison includes CUDA-L1 and KernelSkill \citep{li2026cudal1,sun2026kernelskill}.
It also covers STARK and QiMeng-Kernel \citep{dong2025stark,zhu2026qimeng}, as well as KernelBlaster and AI CUDA Engineer \citep{dong2026kernelblaster,lange2025robustcuda}.
Unless otherwise noted, rows are evaluated on the full KernelBench Level-2 split.
Because existing papers report results with different aggregation conventions, Table~\ref{tab:kernelbench-comparison} keeps arithmetic and geometric means as separate columns while standardizing the most common correctness and fast-$p$ metrics.
The comparison separates correctness, broad speedup, and heavy-tailed arithmetic gains, which are often conflated in aggregate kernel benchmark reports.
LLM4LLM delegates exploit detection in isolated KernelBench runs to the benchmark harness: unconstrained LLMs can produce benchmark-exploiting candidates that hard-code fixed-test outputs or bypass the intended computation.
This is one reason our primary evidence comes from deployment evaluations, where candidates are patched into real model code and accepted through model-integrated correctness and end-to-end latency.

\newcommand{\KernelBenchAblationTable}{%
\begin{table*}[!t]
\centering
\caption{KernelBench Level-2 ablation against PyTorch eager and \texttt{torch.compile}. Pass, Fast1, and Fast2 are reported as percentages; Mean, GeoMean, P50, and P75 are speedups.}
\label{tab:kernelbench-ablation}
\tiny
\setlength{\tabcolsep}{1.25pt}
\renewcommand{\arraystretch}{0.82}
\resizebox{\textwidth}{!}{%
\begin{tabular}{@{}llrrrrrrrrrrrrr@{}}
\toprule
\multicolumn{2}{c}{} &
\multicolumn{7}{c}{\textbf{vs. PyTorch eager}} &
\multicolumn{6}{c}{\textbf{vs. \texttt{torch.compile}}} \\
\cmidrule(lr){3-9}\cmidrule(lr){10-15}
\textbf{Model} & \textbf{Method} & \textbf{Pass} &
\textbf{Fast1} & \textbf{Fast2} & \textbf{Mean} & \textbf{GeoMean} & \textbf{P50} & \textbf{P75} &
\textbf{Fast1} & \textbf{Fast2} & \textbf{Mean} & \textbf{GeoMean} & \textbf{P50} & \textbf{P75} \\
\midrule
\multirow{7}{*}{GPT-5.4}
& Sample-1 & 46 & 4 & 1 & 0.693 & 0.555 & 0.511 & 0.640 & 2 & 1 & 0.643 & 0.507 & 0.500 & 0.562 \\
& Sample-10 & 75 & 63 & 8 & 1.417 & 1.230 & 1.064 & 1.336 & 44 & 9 & 1.344 & 1.124 & 1.009 & 1.323 \\
& Iter-10 & 100 & 64 & 4 & 1.171 & 0.999 & 1.011 & 1.220 & 41 & 5 & 1.065 & 0.887 & 0.980 & 1.144 \\
& LLM4LLM w/o restart (10) & 100 & 89 & 32 & 6.782 & 1.961 & 1.250 & 4.221 & 66 & 31 & 6.573 & 1.742 & 1.143 & 3.986 \\
& LLM4LLM (10) & 100 & 91 & 32 & 14.962 & 2.005 & 1.332 & 4.110 & 71 & 31 & 14.716 & 1.781 & 1.230 & 4.020 \\
& LLM4LLM w/o restart (15) & 100 & 90 & 32 & 7.026 & 1.973 & 1.255 & 4.221 & 67 & 31 & 6.813 & 1.752 & 1.143 & 3.986 \\
\highlightrow & LLM4LLM (15) & 100 & 95 & 35 & 15.929 & 2.153 & 1.437 & 4.691 & 74 & 35 & 15.665 & 1.912 & 1.257 & 4.351 \\
\midrule
\multirow{7}{*}{Claude Sonnet 4.6}
& Sample-1 & 63 & 35 & 4 & 1.663 & 1.079 & 1.009 & 1.235 & 29 & 4 & 1.548 & 0.962 & 0.990 & 1.100 \\
& Sample-10 & 90 & 69 & 9 & 2.117 & 1.302 & 1.097 & 1.414 & 51 & 11 & 2.018 & 1.179 & 1.020 & 1.291 \\
& Iter-10 & 100 & 62 & 4 & 1.183 & 1.063 & 1.036 & 1.342 & 43 & 5 & 1.069 & 0.944 & 0.988 & 1.218 \\
& LLM4LLM w/o restart (10) & 100 & 93 & 35 & 17.346 & 2.267 & 1.497 & 4.903 & 73 & 35 & 17.058 & 2.013 & 1.299 & 4.516 \\
& LLM4LLM (10) & 99 & 98 & 39 & 28.114 & 2.447 & 1.672 & 4.537 & 76 & 39 & 27.747 & 2.171 & 1.379 & 4.542 \\
& LLM4LLM w/o restart (15) & 100 & 93 & 38 & 17.614 & 2.417 & 1.591 & 5.164 & 76 & 38 & 17.317 & 2.146 & 1.436 & 5.117 \\
\highlightrow & LLM4LLM (15) & 100 & 99 & 41 & 28.002 & 2.546 & 1.683 & 5.017 & 77 & 42 & 27.638 & 2.261 & 1.428 & 5.051 \\
\midrule
\multirow{7}{*}{GLM-5}
& Sample-1 & 58 & 37 & 14 & 2.272 & 1.341 & 1.196 & 1.986 & 31 & 17 & 2.191 & 1.219 & 1.022 & 2.333 \\
& Sample-10 & 87 & 76 & 30 & 4.962 & 1.953 & 1.439 & 3.748 & 59 & 33 & 4.813 & 1.751 & 1.286 & 3.594 \\
& Iter-10 & 100 & 70 & 25 & 2.207 & 1.336 & 1.218 & 1.999 & 55 & 26 & 2.091 & 1.187 & 1.049 & 2.081 \\
& LLM4LLM w/o restart (10) & 97 & 86 & 36 & 16.014 & 2.179 & 1.526 & 4.412 & 67 & 35 & 15.696 & 1.935 & 1.303 & 4.328 \\
& LLM4LLM (10) & 100 & 96 & 36 & 20.501 & 2.539 & 1.555 & 5.113 & 76 & 37 & 23.199 & 2.255 & 1.333 & 5.181 \\
& LLM4LLM w/o restart (15) & 99 & 94 & 37 & 16.902 & 2.294 & 1.590 & 4.581 & 71 & 36 & 16.553 & 2.020 & 1.303 & 4.482 \\
\highlightrow & LLM4LLM (15) & 100 & 97 & 39 & 22.157 & 2.745 & 1.680 & 5.539 & 76 & 40 & 24.718 & 2.437 & 1.434 & 5.281 \\
\bottomrule
\end{tabular}}
\end{table*}%
}

\subsection{Ablation Study}
\label{sec:eval-ablation}

We further isolate the contribution of sampling, iterative refinement, and LLM4LLM's restart strategy across GPT-5.4, Claude Sonnet 4.6, and GLM-5.
Each row reports one complete evaluation over the 100 KernelBench Level-2 tasks.
For each task, we apply the candidate budget shown in the method name, retain the fastest correct candidate after KernelBench warmup and repeated timing, and aggregate the resulting task-level speedups using Mean, GeoMean, P50, and P75.
Table~\ref{tab:kernelbench-ablation} reports the same method grid against PyTorch eager and \texttt{torch.compile}.
The first three rows in each model block follow KernelBench-style sampling and execution-feedback iteration \citep{ouyang2025kernelbench}, separating candidate diversity from refinement.
Sample-10 improves substantially over Sample-1, indicating that independent diversity is a strong baseline for finding compilable and occasionally fast kernels.
Iter-10 reaches a 100\% pass rate for all three models, while its GeoMean and Fast2 metrics trail the stronger sampling runs in several settings; the search often spends many turns repairing one trajectory after the first viable implementation.

The LLM4LLM variants add deployment-aware task construction, validation feedback, and episodic restart with compact experience.
With restart, the 15-trial setting gives the best GeoMean for all three models against eager execution (2.153, 2.546, and 2.745) and also improves the \texttt{torch.compile} comparison.
The percentile columns contextualize the heavy-tailed arithmetic means: GLM-5 reaches P50/P75 speedups of 1.680/5.539 against eager and 1.434/5.281 against \texttt{torch.compile}, while Claude Sonnet 4.6 reaches 1.683/5.017 against eager.
Comparing ``w/o restart'' with full LLM4LLM across all three models shows that restart contributes beyond a larger optimization budget, supporting the episodic design in Section~\ref{sec:method-optimization}.
Increasing the budget from 10 to 15 trials yields consistent gains in GeoMean and Fast2, so the 15-trial configuration is used as the strongest KernelBench setting.

\FloatBarrier

\KernelBenchAblationTable

\section{Related Work}
\label{sec:related}

\subsection{Tensor Program and Kernel Optimization}
\label{sec:related-kernel}

Compiler and scheduling systems generate efficient tensor programs across operator and graph levels.
Halide separates algorithms from schedules \citep{ragan2013halide}, while Tensor Comprehensions and TensorIR expose schedule-oriented tensor abstractions \citep{vasilache2018tensor,feng2023tensorir}.
TVM, AutoTVM, and Ansor combine tensor-program generation with learned search and task scheduling \citep{chen2018tvm,chen2018learning,zheng2020ansor}; Triton and OpenTuner cover handwritten GPU programs and extensible autotuning \citep{tillet2019triton,ansel2014opentuner}.
Graph- and model-level systems optimize substitutions, fusion, and dynamic execution across larger computation regions \citep{jia2019taso,jia2019relaxed,ma2020rammer,ansel2024pytorch2}.
LLM4LLM accepts generated kernels through the patched model under a deployment workload.

\subsection{LLM Agents for Code and Kernel Generation}
\label{sec:related-agents}

Large language models have been applied to code synthesis, repair, feedback-driven improvement, and repository-level editing.
Codex, CodeT5, InCoder, and Code Llama establish generation, identifier-aware pretraining, and infilling foundations \citep{chen2021evaluating,wang2021codet5,fried2023incoder,roziere2023code}.
Search and feedback improve programs in competitive programming, repository repair, and self-refinement settings \citep{li2022competition,le2022coderl,jimenez2024swebench,madaan2023self}.
Reflexion, ReAct, self-debugging, and LLM compiler models connect reasoning traces with execution feedback \citep{shinn2023reflexion,yao2023react,chen2024selfdebug,cummins2024llmcompiler}.
KernelBench focuses this capability on GPU kernels \citep{ouyang2025kernelbench}; recent systems add reinforcement learning and multi-agent planning \citep{li2026cudal1,sun2026kernelskill,dong2025stark,zhu2026qimeng}.
Memory and verification further address cross-task reuse and benchmark reliability \citep{dong2026kernelblaster,lange2025robustcuda}.
LLM4LLM places the search loop inside a deployment-aware pipeline conditioned on the target inference script.

\subsection{Language Model Inference and Serving Optimization}
\label{sec:related-serving}

Language-model serving systems optimize memory, batching, scheduling, and execution phases.
ORCA and vLLM target iteration scheduling and KV-cache management \citep{yu2022orca,kwon2023efficient}, while Sarathi and SGLang study chunked execution and structured generation runtimes \citep{agrawal2023sarathi,zheng2023sglang}.
FlexGen, DeepSpeed Inference, and DeepSpeed-FastGen address memory- and throughput-oriented generation at larger scales \citep{sheng2023flexgen,aminabadi2022deepspeed,holmes2024deepspeedfastgen}.
At the kernel level, FlashAttention/FlashAttention-2 and FlashInfer provide optimized attention paths for inference workloads \citep{dao2022flashattention,dao2023flashattention,ye2025flashinfer}.
State-space and hybrid recurrent models add fast paths through Mamba, Mamba-2, causal depthwise convolution, and RecurrentGemma-style recurrent blocks \citep{gu2023mamba,mamba2,dao2024causalconv1d,botev2024recurrentgemma}.
LLM4LLM uses this deployment context as the environment in which generated candidates are validated and accepted.

\section{Conclusion}
\label{sec:conclusion}

LLM4LLM reframes LLM-based kernel optimization as a deployment-aware closed-loop problem.
The framework connects real-model profiling, phase-aware task construction, experience-guided episodic search, and deployment-time acceptance.
By making the target workload part of candidate generation and acceptance, LLM4LLM aligns kernel search with the execution context that determines inference performance.
Across diverse language-model families and two GPU platforms, this loop converts generated candidates into end-to-end inference gains while clarifying the relationship between isolated benchmark performance and deployment behavior.
The results suggest a practical path for deployable LLM-generated kernels: profile the real model, search under phase-aware constraints, and accept candidates through target-runtime validation.


\section*{Limitations}

LLM4LLM currently targets single-GPU inference workloads and assumes access to a representative inference script.
The system studies deployment-aware optimization at the module and model-instance level; tensor parallelism, pipeline parallelism, continuous batching, and multi-tenant serving introduce additional acceptance criteria, including communication cost, scheduler interaction, and batch-level interference.
The method specializes patches to observed shape and phase regimes, so deployment settings with substantially different prompts, decode lengths, batching behavior, or model configurations require re-profiling and renewed acceptance checks.
Optimization cost is also a practical consideration: search is most attractive when the resulting patch is reused across many inference calls or related model instances.
The quality of generated candidates depends on the coding ability of the underlying model, the search budget, and the availability of relevant implementation patterns.
Future work can extend deployment-time acceptance to distributed serving systems and richer workload mixtures.

\section*{Acknowledgments}

We thank the anonymous reviewers and the meta-reviewer for their constructive feedback.  This work was supported by the Zhongguancun Academy (Grant No. C20250501). This work was also supported in part by the Shaanxi Key Technology R\&D Program under Grant 2024GX-ZDCYL-02-15, in part by the Natural Science Funds for Distinguished Young Scholar of Shaanxi under Grant 2025JC-JCQN-079. This work was also supported by the National Key Laboratory of Advanced Communication Networks (Grant No. FFX26641X006).

\bibliography{custom}

\clearpage
\appendix
\setcounter{table}{0}
\renewcommand{\thetable}{\Alph{section}.\arabic{table}}

\section{Appendix}
\label{sec:appendix-additional-analysis}

This appendix expands on mechanisms that are only summarized in the main paper.
The examples come from saved run artifacts and use the same acceptance principle as the main experiments: a candidate is timed only after it passes the corresponding correctness check.
Section~\ref{sec:appendix-restart-cases} analyzes restart as search-state control; Section~\ref{sec:appendix-sampling-iteration} isolates search diversity from linear repair; Section~\ref{sec:appendix-api-fix} discusses API compatibility for fast-moving kernel DSLs; and Sections~\ref{sec:appendix-model-differences}--\ref{sec:appendix-core-kernels} summarize model-backbone behavior and accepted deployment kernels.

\subsection{Search-State Control by Restart}
\label{sec:appendix-restart-cases}

Table~\ref{tab:appendix-restart-case} shows a restart case from KernelBench Level 2, problem 97, \texttt{Matmul\_BatchNorm\_BiasAdd\_Divide\_Swish}.
Both variants use GPT-5.4, Triton fp32, an A100, and a 15-candidate budget.
The no-restart run first found a correct epilogue-only candidate, then repeated the same output-mismatch signature for three turns, and eventually returned to a family of correct epilogue-only implementations around 8 ms.
The restart-enabled run produced full \texttt{Linear + BatchNorm + bias/divide/Swish} fusion inside the GEMM epilogue, reducing the best latency from 8.01 ms to 1.42 ms.

\begin{center}
\small
\setlength{\tabcolsep}{5pt}
\renewcommand{\arraystretch}{1.08}
{\captionsetup{hypcap=false}\captionof{table}{Restart case on KernelBench Level 2 problem 97, measured on A100. The baseline eager runtime is 8.23 ms for both rows.}\label{tab:appendix-restart-case}}
\begin{tabular}{lrr}
\toprule
\textbf{Variant} & \textbf{Best ms} & \textbf{Speedup} \\
\midrule
No restart & 8.01 & 1.03$\times$ \\
Restart & 1.42 & 5.80$\times$ \\
\bottomrule
\end{tabular}
\end{center}

This case illustrates the role of restart as search-state control.
The useful feedback is compact: the operation sequence, the inference-mode BatchNorm contract, and the need to preserve the bias and activation semantics.
The long repair history contains many local choices tied to one partially fused layout.
Restart carries forward the compact constraints while allowing the next episode to reselect the optimization scope, which can move the search from epilogue fusion to GEMM-level fusion.
The generated code exposes the difference directly:
\begin{quote}
\scriptsize
\begin{verbatim}
# no restart: best candidate
x = self.matmul(x)
x = self.bn(x)
x = fused_bias_div_swish(
    x, self.bias, self.divide_value)

# restart: best candidate
return fused_linear_bn_bias_div_swish(
    x, self.matmul.weight, self.matmul.bias,
    self.bn.running_mean, self.bn.running_var,
    self.bn.weight, self.bn.bias, self.bias,
    self.bn.eps, self.divide_value)
\end{verbatim}
\end{quote}
The corresponding kernel bodies show the same distinction:
\begin{quote}
\scriptsize
\begin{verbatim}
# no restart: epilogue-only kernel
y = (x + b) * inv_divide_value
y = y * tl.sigmoid(y)

# restart: GEMM-level fused kernel
acc += tl.dot(x, w)
acc = acc + lin_b[None, :]
acc = (acc - mean[None, :]) * inv_std[None, :]
acc = acc * gamma[None, :] + beta[None, :]
acc = (acc + extra_b[None, :]) / divide_value
acc = acc * tl.sigmoid(acc)
\end{verbatim}
\end{quote}
In the no-restart trace, iteration 1 is correct at 8.37 ms, iterations 2--4 repeat the same output-mismatch signature, and iterations 5--15 return to correct epilogue-only variants around 8 ms.
With restart enabled, correct full-fusion candidates appear at iteration 2 (1.45 ms) and iteration 3 (1.42 ms), while later candidates explore both fast and slow alternatives.

\subsection{Sampling and Iteration Are Complementary}
\label{sec:appendix-sampling-iteration}

Table~\ref{tab:appendix-sample-iter-case} gives two complementary contrasts from GPT-5.4 KernelBench Level 2 runs.
Problem 76, \texttt{Gemm\_Add\_ReLU}, shows the value of search diversity: sampling found a full-fusion implementation, while the recorded iterative run stopped at a correct candidate that left GEMM in PyTorch and fused only the bias-ReLU epilogue.
Problem 1, \texttt{Conv2D\_ReLU\_BiasAdd}, shows the reverse pattern: the best one-shot sample was correct but slower than eager execution, while iterative repair fixed an initial output mismatch and produced a faster NCHW-specialized epilogue kernel.
The point is not that either sampling or iteration dominates; the two mechanisms expose different useful candidates.

\begin{center}
\small
\setlength{\tabcolsep}{4pt}
\renewcommand{\arraystretch}{1.08}
{\captionsetup{hypcap=false}\captionof{table}{Complementary sampling and iteration examples from GPT-5.4 KernelBench Level 2 runs on A100. Speedup is against the eager baseline for each problem.}\label{tab:appendix-sample-iter-case}}
\begin{tabular}{llrr}
\toprule
\textbf{Problem} & \textbf{Method} & \textbf{Best ms} & \textbf{Speedup} \\
\midrule
76 & Sample-10 & 1.39 & 5.76$\times$ \\
76 & Iter-10 & 8.11 & 0.99$\times$ \\
1 & Sample-10 & 8.12 & 0.90$\times$ \\
1 & Iter-10 & 6.02 & 1.21$\times$ \\
\bottomrule
\end{tabular}
\end{center}

For problem 76, the core generated code shows the optimization-scope distinction.
The sampled candidate fuses the matrix multiplication and epilogue into one Triton implementation:
\begin{quote}
\scriptsize
\begin{verbatim}
# Sample-10 best candidate
return fused_linear_bias_relu(x, weight, bias)

# kernel core
acc += tl.dot(a, b)
acc += bias[None, :]
acc = tl.maximum(acc, 0.0)
\end{verbatim}
\end{quote}
The iterative candidate is correct, but its generated structure preserves the PyTorch GEMM and only moves the epilogue to Triton:
\begin{quote}
\scriptsize
\begin{verbatim}
# Iter-10 best candidate
y = self.gemm(x)
y = triton_bias_relu(y, self.bias)
return y

# kernel core
y = tl.maximum(x + b[None, :], 0.0)
\end{verbatim}
\end{quote}
For problem 1, the useful signal is different.
The iterative run first produced an output mismatch; after one repair turn, it kept the same conservative module boundary as the sampled code but generated a cleaner NCHW-specialized epilogue kernel:
\begin{quote}
\scriptsize
\begin{verbatim}
# Sample-10 best candidate
x = self.conv(x)
x = triton_relu_bias(x, self.bias)
return x

# kernel core
hw = H * W
c = (offs // hw) % C
b = tl.load(bias_ptr + c, mask=mask, other=0.0)
y = tl.maximum(x, 0.0) + b
\end{verbatim}
\end{quote}
\begin{quote}
\scriptsize
\begin{verbatim}
# Iter-10 best candidate, after repair
x = self.conv(x)
x = triton_relu_bias_nchw(x, self.bias)
return x

# kernel core
c = (offs // HW) % C
b = tl.load(bias_ptr + c, mask=mask, other=0.0)
y = tl.maximum(x, 0.0) + b
\end{verbatim}
\end{quote}
Together, these examples explain why LLM4LLM combines sampling, feedback, and restart instead of treating iterative repair as a purely monotonic process.
Sampling exposes alternative decompositions of the same PyTorch graph, while linear repair is effective for turning a nearby candidate into a valid and better-specialized one.
Restart combines these roles: each new episode can choose a fresh optimization scope, but it still receives compact correctness and implementation constraints learned from previous attempts.

\subsection{API Compatibility and User-Defined Fixes}
\label{sec:appendix-api-fix}

This section separates interface compatibility from the optimization-scope issue in Section~\ref{sec:appendix-sampling-iteration}.
Kernel DSLs such as Triton and TileLang evolve quickly, and local installations can differ from the APIs seen during model training.
In KernelBench Level 2 problem 86, \texttt{Matmul\_Divide\_GELU}, an iterative candidate failed because the generated GELU approximation called an unavailable Triton math entry point:
\begin{quote}
\scriptsize
\begin{verbatim}
# failed iterative candidate
inner = c0 * (x + c1 * x * x * x)
y = 0.5 * x * (1.0 + tl.math.tanh(inner))
\end{verbatim}
\end{quote}
After repair, the run obtained a correct epilogue-only kernel, while an independent sampled candidate reached a fused \texttt{Linear + divide + GELU} implementation:
\begin{center}
\small
\setlength{\tabcolsep}{4pt}
\renewcommand{\arraystretch}{1.08}
{\captionsetup{hypcap=false}\captionof{table}{API-affected example on KernelBench Level 2 problem 86, measured on A100.}\label{tab:appendix-api-case}}
\begin{tabular}{lrr}
\toprule
\textbf{Method} & \textbf{Best ms} & \textbf{Speedup} \\
\midrule
Sample-10 & 2.85 & 2.82$\times$ \\
Iter-10 & 8.01 & 1.00$\times$ \\
\bottomrule
\end{tabular}
\end{center}
The generated code differs at the kernel boundary:
\begin{quote}
\scriptsize
\begin{verbatim}
# Sample-10 best candidate
return triton_linear_div_gelu(
    x, self.linear.weight,
    self.linear.bias, self.divisor)

# Iter-10 repaired candidate
x = self.linear(x)
x = triton_div_gelu(x, self.divisor)
return x
\end{verbatim}
\end{quote}
LLM4LLM therefore includes an API-correction layer that users can extend for their local backend versions.
The correction file is backend specific and currently covers Triton, TileLang, and CUDA extension patterns, for example:
\begin{quote}
\scriptsize
\begin{verbatim}
tl.math.tanh  -> libdevice.tanh
tl.math.max   -> tl.maximum
tl.math.min   -> tl.minimum
T.Ranged      -> tilelang.language.Range
data<T>()     -> data_ptr<T>()
\end{verbatim}
\end{quote}
These fixes keep version-dependent interface repair separate from performance-relevant kernel design decisions such as fusion scope, tiling, masking, accumulator precision, cache updates, and launch structure.

\subsection{Behavior Across LLM Backbones}
\label{sec:appendix-model-differences}

The ablation table shows that LLM backbones differ in more than final pass rate.
GPT-5.4 follows repair feedback reliably, but the examples above show that it can become conservative once a correct partial-fusion implementation is available.
Restart and experience summaries are useful in this setting because they preserve stable constraints without preserving every local edit in the failed trajectory.

Claude Sonnet 4.6 tends to produce well-guarded code with explicit fallback paths.
This behavior helps coverage and makes deployment-time acceptance easier to apply, while the search loop still has to test whether guarded candidates enter the optimized path under real prefill/decode conditions.
GLM-5 produces more aggressive candidates and higher upper-tail speedups in the KernelBench ablation.
The same search loop accommodates these behaviors by using identical correctness, timing, and deployment-acceptance rules for all backbones.

\FloatBarrier
\subsection{Deployment Kernel Families}
\label{sec:appendix-core-kernels}

Table~\ref{tab:appendix-core-kernel-cases} summarizes representative accepted kernels from extracted language-model tasks.
The numbers are task-level validation speedups on extracted modules; the end-to-end effect after patching is reported in Table~\ref{tab:main-e2e} and Figures~\ref{fig:attention-scope-compare}--\ref{fig:mamba-scope-compare}.

\begin{table*}[t]
\centering
\small
\setlength{\tabcolsep}{4pt}
\renewcommand{\arraystretch}{1.08}
\caption{Representative accepted Triton kernels on extracted deployment tasks, measured on A100. Speedup is measured on the extracted validation task for the module, before full-model aggregation.}
\label{tab:appendix-core-kernel-cases}
\resizebox{\textwidth}{!}{%
\begin{tabular}{llllr}
\toprule
\textbf{Family} & \textbf{Module} & \textbf{Phase / bucket} & \textbf{Main specialization} & \textbf{Task speedup} \\
\midrule
Transformer & \texttt{LlamaAttentionPrefill} & \texttt{q\_len > 1} & Causal online-softmax attention with fused QKV wiring & 2.40$\times$ \\
Transformer & \texttt{Qwen3AttentionDecode} & \texttt{q\_len = 1} & Grouped-query decode sharing each KV tile across query heads & 2.01$\times$ \\
Mamba2 & \texttt{Mamba2MixerPrefill} & prefill scan & Compact recurrent scan with cache-safe mixer wiring & 14.31$\times$ \\
Mamba & \texttt{MambaMixerDecode} & single-token decode & Fused Conv1d--SiLU--projection chain for cached decode & 2.06$\times$ \\
Recurrent & \texttt{RecurrentGemmaRglru} & recurrent update & Fused gate normalization plus recurrent-state scan & 6.85$\times$ \\
\bottomrule
\end{tabular}}
\end{table*}

\noindent\textbf{Transformer attention.}
The accepted attention kernels are phase specialized.
For prefill, the kernel maps one Triton program to a query-head and query-token tile, computes the KV head from the grouped-query structure, embeds causal masking in the score update, and maintains online-softmax statistics in fp32.
This structure reduces memory traffic by keeping the attention matrix implicit and by sharing KV loads according to the model's head grouping.
For decode, the kernel targets the single-token regime and processes one KV head per program while computing the associated query-head group together.
The decode path also keeps mask handling and output layout aligned with the patched module, so the surrounding projection and cache update see the same tensor contract as eager execution.

\noindent\textbf{Mamba-family state-space mixers.}
The Mamba-family cases show the value of optimizing a deployable module boundary that spans multiple primitive operations.
For Mamba2 prefill, the accepted path preserves the mixer sequence of input projection, depthwise convolution, split hidden/state parameters, recurrent scan, gated RMSNorm, and output projection.
Inside the recurrent kernel, compact \(B\) and \(C\) state vectors are loaded once per time step and broadcast logically across heads, while the state tile remains in registers across the scan.
For decode, the accepted kernels specialize to \texttt{seq\_len = 1} with an existing cache, update convolution and SSM state explicitly, and minimize launch overhead for the recurring single-token path.
These cases explain why profiling selects Mamba mixers as high-impact targets in Table~\ref{tab:main-e2e}.

\noindent\textbf{RecurrentGemma RGLRU.}
RecurrentGemma exposes a gated recurrent core whose latency comes from both elementwise gate transformations and recurrent state movement.
The accepted RGLRU implementation fuses sigmoid, softplus-derived recurrent gating, reset handling, normalization, and the recurrent scan into Triton kernels that preserve the state update semantics.
The key advantage is that the recurrent state is carried through the sequence inside the kernel and the final state is written once, reducing Python-level dispatch and intermediate tensor traffic.
This case is distinct from attention and Mamba: the dominant optimization target is the recurrent update itself, which supports the profiling-first design used by LLM4LLM.

Across these families, the accepted kernels are useful because the generated code respects the module boundary that the deployment patch will actually replace.
For attention, this means that the code must match the phase-local cache contract: a prefill kernel can assume a query block and construct causal tiles, while a decode kernel must read a populated KV cache and update only the single-token output path.
For Mamba and recurrent blocks, the same principle appears as state ownership rather than KV ownership.
The kernel must update convolution, SSM, or recurrent state exactly once and return tensors with the same layout expected by the surrounding model.
These examples are therefore not just faster isolated kernels.
They are accepted because the optimized path can be inserted into the profiled model instance without changing the caller-visible tensor, cache, or phase semantics.

\subsection{Summary}
\label{sec:appendix-case-takeaways}

The concrete run traces support the design choices in the main paper.
Restart is useful when a short set of constraints should survive but a long repair trajectory should not dominate the next implementation.
Sampling provides optimization-scope diversity that a single linear repair path may not expose.
API fixes separate version-dependent DSL compatibility from performance-relevant kernel design.
The deployment cases show that the accepted kernels are not one generic template: attention, Mamba mixers, and recurrent blocks require different phase contracts and cache semantics.

\end{document}